\documentclass[conference]{IEEEtran}
\usepackage{cite}

\ifCLASSINFOpdf
  \usepackage[pdftex]{graphicx}
\else
  \usepackage[dvips]{graphicx}
\fi
\usepackage{amsmath}
\usepackage{tabularx,booktabs}
\newcolumntype{C}{>{\centering\arraybackslash}X}
\usepackage{array}
\usepackage{url}
\usepackage{hyperref}

\usepackage{color}
\usepackage{xcolor}
\newcommand{\minisection}[1]{\vspace{0.03in} \noindent {\bf #1}}

\usepackage{todonotes}
\newcommand\plotwidth{0.31}

\begin{document}
%
\title{
Policy Fusion for Adaptive and Customizable Reinforcement Learning Agents
}

\author{
\IEEEauthorblockN{Alessandro Sestini}
\IEEEauthorblockA{\textit{Dipartimento di Ingegneria} \textit{dell'Informazione},\\
\textit{Università degli Studi di Firenze},\\
Florence, Italy\\
alessandro.sestini@unifi.it}
\and
\IEEEauthorblockN{Alexander Kuhnle}
\IEEEauthorblockA{\textit{University of Cambridge,}\\
Cambridge, United Kingdom\\
alexander.kuhnle@cantab.net}
\and
\IEEEauthorblockN{Andrew D. Bagdanov}
\IEEEauthorblockA{\textit{Dipartimento di Ingegneria} \textit{dell'Informazione},\\
\textit{Università degli Studi di Firenze},\\
Florence, Italy\\
andrew.bagdanov@unifi.it}}


%


\maketitle

\begin{abstract}
  In this article we study the problem of training intelligent agents using Reinforcement Learning for the purpose of game development. Unlike systems built to replace
  human players and to achieve super-human performance, our agents aim to produce meaningful
  interactions with the player, and at the same time demonstrate behavioral traits as desired
  by game designers.
  We show how to combine distinct
  behavioral policies to obtain a meaningful ``fusion'' policy which comprises all these behaviors. To this end, we propose four different policy fusion methods for combining pre-trained policies. We further
  demonstrate how these methods can be used in combination with Inverse
  Reinforcement Learning in order to create intelligent agents with specific behavioral styles as chosen by game designers, without having to
  define many and possibly poorly-designed reward functions. Experiments on two
  different environments indicate that entropy-weighted policy fusion
  significantly outperforms all others. We provide several practical examples
  and use-cases for how these methods are indeed useful for video game production
  and designers.
\end{abstract}


%
\IEEEpeerreviewmaketitle
\section{Introduction}
\label{sec:intro}
Non-player characters (NPCs) are a vital factor with respect to the quality of a video
game, with the potential to elevate or ruin the player experience. Recent
advances in Deep Reinforcement Learning (DRL) have managed to emulate and even supersede human players in the pursuit of achieving (super-)human-level performance, and
examples of NPC agents trained with such techniques have been demonstrated for
commercial video games \cite{sourcemadness, motogp}. 

However, mass adoption of DRL by game designers still needs significant technical
innovation to make them compatible with the realities of the modern game development process, 
and to build trust in these approaches~\cite{microsoft}. 
Game AI designers need to be able to meaningfully guide their exploration of the
space of possible behaviors via quick design iterations, without having to tune
opaque hyperparameters and study training analytics which are hard to interpret with respect 
to agent behavior. 
We are interested in training NPCs that exhibit specific behavioral styles or
attitudes \emph{chosen by designers}. However, achieving this using standard DRL
training -- i.e. via definition and shaping of a complex reward function -- can
be impractical \cite{safety} and expensive since it can require hundreds of
thousands of episodes to learn useful policies from scratch. Techniques like
Imitation Learning (IL)~\cite{bc1} and Inverse Reinforcement Learning
(IRL)~\cite{russelng} can help game designers on the first point, by providing
tools to articulate behaviors without requiring handcrafted rewards, however
they do not address the sample efficiency problem inherent in DRL training. The
policy fusion approaches we propose in this paper pair well in particular with
Inverse Reinforcement Learning in that it is possible to train self-contained,
\emph{micro-behaviors} from expert (designer) demonstrations that are then
\emph{fused} with the main agent policy to adapt it with the new behavior.


\begin{figure}
    \centering
    \includegraphics[width=0.33\textwidth]{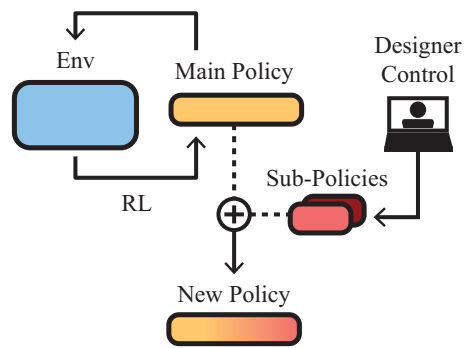} 
    \caption{Summary of our approach: by combining a previously-trained \emph{main policy} with with one or more
      {sub-policies}, we can add local behaviors to the overall agent behavior,
      adapting it to evolving game design instead of discarding the previously trained agent policy and training a novel one from scratch. In this way, designers can create agents
      that reflect specific choices in easy and understandable ways.}
    \label{fig:teasing}
\end{figure}

We identify three situations that designers face when training game agents
with DRL. In all of the following, we assume the existence of an agent trained with standard DRL which, however, does not reflect designer intent
-- and which therefore requires adaptation to:
\begin{itemize}
\item \textbf{Enhance performance}, when the agent trained with standard RL does not
  meet performance requirements -- often due to a sparse or poorly crafted reward function. In standard RL,
  addressing this usually means tuning hyperparameters, modifying the training setup, adjusting the reward function, and then 
  re-training the agent;
\item \textbf{Add style}, when the agent does not reflect the qualitative
  behavior desired by designers. For example, an agent might be supposed to act
  more ``sneaky'' or more ``aggressive'' relative to the current agent behavior. In standard RL, this usually means adapting the reward function accordingly and
  re-train the agent; and
\item \textbf{Adapt to new features}, when designers change a detail in the game mechanics such as adding a new object/ability/property/etc.
  Again, this typically requires re-training a previously trained agent from
  scratch. Moreover, one cannot rely on the fact that the previous training
  procedure and hyperparameters will continue to work well in this new version of the
  environment.
\end{itemize}

In this paper we tackle the aforementioned problems and propose methods to
combine diversified policies in ways that avoid re-training. As illustrated in
figure~\ref{fig:teasing}, instead of discarding the previously trained agent policy and training a novel one from scratch, we
instead train a sub-policy
to handle the intended behaviour aspect. For example, a sub-policy may be trained only to learn how to
handle a novel object added by a designer. We then merge the main policy with the
sub-policy via policy fusion, which requires no re-training and ideally results
in an agent able to properly handle this new object while maintaining the overall skills of the main game policy. We also demonstrate how these approaches can
efficiently be used in combination with IRL in order to train agents which reflect the intention of designers without requiring hand-crafted reward shaping. Experiments on two game environments show that our policy fusion approaches
outperform fusion methods from the literature
and that fused policies achieve the
same -- in some cases even better -- results than re-training the main policy from scratch using
engineered reward functions.

\section{Related work}
The potential of DRL in video games has been gaining interest from the research
community. Here we review recent work most related to our contributions.

\minisection{DRL \emph{in} video games.} Modern video games are environments
with complex dynamics which can be useful testbeds for complex DRL algorithms.
Results from AlphaStar \cite{alphastar} and OpenAI 5 \cite{openai2019dota}
demonstrated how DRL can be used to create super-human agents in modern
complex video-games, while results such as Ecoffet et al.\ \cite{goexplore} show
that we can create super-human agents able to surpass human players in the ATARI
games of the Arcade Learning Environment (ALE) \cite{ale}. However, our motivation is different in that we do not aim to create super-human agents, but rather to facilitate the use of DRL for NPCs as part of 
game design.

\minisection{DRL \emph{for} video games.} At the same time, there is an
increasing interest from the game development community in how to properly use
DRL for video game production. As stated in the introduction, Jacob et al.\ \cite{microsoft} argued that industry does not need agents build to ``beat the
game'', but rather to produce credible and human-like behaviors. Results such as
\cite{ubisoft, ubisoftnavigation, unityaction} are other notable examples of
applying DRL to commercial video games, as well as the DeepCrawl work 
\cite{sesto19, transformers}, a game environment specicially designed for
studying the applicability of DRL in video game production. 
Within this context, Procedural Content
Generation (PCG) has recently gained a lot of attention: works like
\cite{procenv, procedural, illuminating} have noticed that diverse environment
distributions are essential to adequately train and evaluate RL agents for video
game production, as these kinds of environments enable generalization of agents
when faced with design changes \cite{transformers}.

\minisection{Inverse Reinforcement Learning.}
IRL refers to techniques
that infer a reward function from human demonstrations, which can
subsequently be used to train an RL policy. Adversarial Inverse
Reinforcement Learning (AIRL) \cite{airl} is a
state-of-the-art IRL method, which is enhanced in \cite{deairl} to work with
PCG environments.  Other IRL methods different from AIRL, which were tested on simple and static
environments, are \cite{trex, pref,
  prefplusimi}. There are only few examples of applying IRL for video
game agents: Tucker et al. \cite{adam} try to use AIRL on ALE without
success, while Source of Madness \cite{sourcemadness} used Imitation
Learning to create diverse 
game agents in a commercial video game.

\minisection{Ensemble methods for RL.} The use of ensemble methods in
RL refers to the practice of combining two or more RL algorithms to
increase their performance. Wiering et
al.\ \cite{ensemble}
survey different methods to combine
multiple RL algorithms. Other examples of ensemble
methods in RL are \cite{ensembleapprox, ensemblerobust,
  ensembleterrain, ensemblevehicle, mult2}. 
However, all of these techniques
combine models but with the same goal, instead our aim is to combine
policies with different objectives, possibly even orthogonal ones.

Multi-objective learning is motivated by a similar goal, and attempts to
combine different reward functions during training to create a
complex agent \cite{multisurvey, multiaction, multiconfidence,
  multishaping}.  Concurrently to our work, Aytemiz et
al.\ \cite{microsoftblog} started to tackle a similar problem to ours with a
multi-objective approach. These approaches, however, try to combine different
objectives during training. Instead, our aim is to combine various policies 
without re-training of agents.

The \emph{policy fusion} approaches we describe in this paper are distinct from
ensemble methods and multi-objective learning. Our objective is to combine
distributions of different policies after training, while ensemble methods
combine decisions and multi-objective learning combine reward functions.

\section{Background}
IRL plays a critical role for training the sub-policies in our fusion approach. 
Applying IRL to video games and particularly PCG ones is difficult due dependence 
of IRL approaches on the number of demonstrations. 
Demonstration-Efficient Adversarial Inverse
Reinforcement Learning (DE-AIRL) \cite{deairl} addresses this problems and makes it possible to efficiently apply IRL to PCG. This technique is
a modification of the Adversarial Inverse Reinforcement Learning
(AIRL) \cite{airl} algorithm, a GAN \cite{gan} approach that
alternates between training a discriminator $D_\theta(s,a)$ to
distinguish between policy and expert trajectories and optimizing the
trajectory-generating policy $\pi(a|s)$.  The AIRL discriminator is
given by:
\begin{equation}
	\label{eq:discriminator}
	D_\theta(s,a) =
	\frac{\exp\{f_{\theta,\omega}(s,a,s')\}}{\exp\{f_{\theta,\omega}(s,a,s')\} + \pi(a | s)},
\end{equation}
where $\pi(a|s)$ is the generator policy and
$f_{\theta,\omega}(s,a,s') = r_\theta(s,a) + \gamma \phi_\omega(s') -
\phi_\omega(s)$ is a potential base reward function which combines a
reward function approximator $r(s,a)$ and a reward shaping term
$\phi_{\omega}$.  The objective of the discriminator is to minimize
the cross-entropy between expert demonstrations $\tau^E = (s_0^E,
a_0^E, \dots)$ and generated trajectories $\tau^\pi = (s_0^\pi,
a_0^\pi, \dots)$, while the policy is optimized with respect to the
learnt reward function:
\begin{equation}
	\label{eq:rew_discrim}
	\hat{r}(s,a) = \log(D_\theta(s,a)) - \log(1 - D_\theta(s,a)).
\end{equation}


Intuitively, given a state we train the discriminator to recognize if this state
comes from the expert dataset or the policy generator, while the aim of the
policy generator is to fool the discriminator. The best way for the policy to
fool the discriminator is to mimic as well as it can the demonstrations of the
expert dataset, making it hard for the discriminator to recognize if a state
comes from the dataset or the generator. In the end, the discriminator
represents a good reward function for training a near-expert policy with DRL.

DE-AIRL is a modification of the training procedure of AIRL. First, it uses
three extension to the original algorithm in order to increase stability:
\begin{itemize}
\item \textbf{Reward standardization}: the learnt reward model is
  standardized to have zero mean and a defined standard deviation;
\item \textbf{Policy dataset expansion}: instead of using one
  discriminator training step followed by one policy optimization
  step, as the original algorithm, DE-AIRL performs $K$ iterations of
  forward RL for every discriminator step; and
\item \textbf{Fixed number of timesteps}: the terminal
  condition of the environment is removed, because these conditions encode
  information about the environment even when the reward function
  is not observable.
\end{itemize}

Moreover, to reduce the number of demonstrations needed when working with PCG
environments, DE-AIRL introduces an artificially reduced environment, called
SeedEnv, that consists of $n$ levels sampled from the fully procedural
environment, called ProcEnv. The levels sampled are then used to obtain $n$
expert demonstrations:
\begin{gather*}
  \mbox{SeedEnv}(n) = \{L_1, \ldots, L_n \mid L_i \sim \mbox{ProcEnv}\}  \\
  \text{Demos} = \{\tau^{L_i} \mid L_i \in \mbox{SeedEnv}(n) \}.
\end{gather*}
The reward function is trained via AIRL on the reduced SeedEnv environment
instead of the fully-procedural one. In doing so, DE-AIRL forces the
discriminator to focus on expert behavior instead of overfitting to levels
characteristics, substantially reducing the number of expert demonstrations
needed. A reward model trained in this way is able to generalize beyond the
SeedEnv and can thus be used to train an agent on the original environment.

\section{Policy Fusion Methods}


\label{sec:prop_methods}
\begin{figure}
    \centering
    \includegraphics[width=0.45\textwidth]{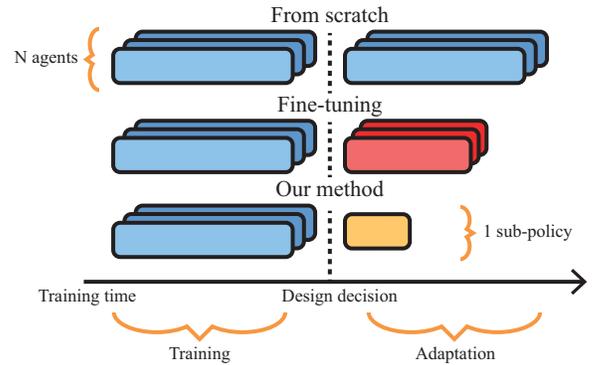} 
	\caption{Training time for different adaptation methods. The
          easiest approach is to simply re-train all agents from
          scratch, but this also requires the most time and assumes
          the training procedure and hyperparameters remain valid
          after design changes.  Instead, with our policy fusion methods we need to
          only train a single sub-policy without touching the already-trained
          agents, requiring about a tenth
          of the time compared to re-training (see section
          \ref{sec:training_times} for more details).}
	\label{fig:training_times}
\end{figure}

When game designers want to create game agents with DRL for commercial
video games, they must face a large amount of challenges.  For
example, they must go through many iterative design choices that
change the environment and force agents to adapt with it
\cite{elettronicarts}; or they need to adapt the final agent
behavior because it does not reflect the designers intention
\cite{microsoft}; or they must quickly test if a new feature
added to the game can work with all the previous changes.

Suppose designers have an agent previously trained with DRL and it must 
be adapted to a game change.
The simplest but most expensive solution is to change something in the
training set-up -- like reward function or environmental dynamics --
and restart the training from scratch. Suppose that to train an agent
from scratch will take $T$ hours.  Usually, a video game does not only have
one agent, but multiple ones with different behaviors
or characteristics. Hence, if we suppose that there are $N$ agents,
training all of them from scratch will take $N \times T$ hours every
time a design decision is made.

Another possibility is to fine-tune the agents every time these
decisions are made. To fine-tune an agent we need to change something
in the training set-up and continue the DRL training of the previously-trained
agent. Therefore, we must spend $t_{ft}$
hours, with generally $t_{ft} \leq T$. However, since there are $N$ agents we need
to fine-tune all of them, resulting in a total of $N \times t_{ft}$
hours.  Moreover, fine-tuning an agent after a design decision is not
always trivial. It is known that DRL suffers from overfitting
\cite{illuminating} that can render the process of
fine-tuning very difficult. However, this can be mitigated with PCG
\cite{transformers}. Finally, both fine-tuning and training from
scratch often require hyper-parameter tuning which may increase the
overall training time.

Our objective is to reduce training time and to avoid the re-training
of all agents every time we make a design decision. Our main idea is
to train sub-policies that explain locally some type of behavior that
designers want to teach to agents. Then, we need a policy fusion
method that can combine the main policy
with the new sub-policy without loosing the skills from its main training. In this way, we get an agent
that is now able to adapt to the new behavior. For example, suppose
we have a well-trained agent but we add a new usable object to the
game, which the agent never saw during training and therefore is
unable to use it. Designers could now train a sub-policy which learns
only how to best use that object. Then, they can combine the
previously trained policy with the new sub-policy in order to ``teach'' the agent
to properly use the new feature.

Suppose we spend $t_{sp}$ hours to train the sub-policy, then most
likely $t_{sp} \leq t_{ft} \leq T$. This method is completely
independent from the number $N$ of previously trained agents, because
we need to only train 1 sub-policy and combine it with any of the
previously trained agents. So, the total training time in this case is
just $t_{sp}$.
Figure \ref{fig:training_times} shows an example of the
different training times.



\label{sec:methods}
We propose four
different policy fusion methods for DRL to combine different policies. The first
two are simple approaches often used for ensemble methods in RL \cite{ensemble,
  mult2}, while the other two are novel techniques proposed by us in this paper.

Suppose we have a main policy $\pi_0$ and a set of sub-policies $\pi_k$ for
$k=1,\ldots,K$. Each policy takes as input a state $s_t$ and returns a
probability distribution over the same discrete action-space. As is common in
the reinforcement learning literature, we will write the policy as
$\pi_i (a | s_t)$ to indicate that it is a distribution over actions conditioned
on state $s_t$. 

To adapt the main policy to include the behaviors of the sub-policies, without
the need for retraining, we propose the following fusions methods which result
in a fusion policy $\pi_f$:
\begin{itemize}
\item \textbf{Mixture Policy (MP)}: the resulting policy
  is the average of the main and all sub-policies:
  \begin{equation}
    \pi_{f}(a | s_t) = \frac{1}{K + 1} \sum^K_{k=0}{\pi_k(a | s_t)}.
  \end{equation}
  
\item \textbf{Product Policy (PP)}: the resulting policy
  is the product of the main and all sub-policies:
  \begin{equation}
    \pi_{f}(a|s_t) = \frac{1}{Z} \prod^K_{k=0}{\pi_k(a|s_t)},
  \end{equation}
  where $Z$ is the normalization constant required to make $\pi_f$ a probability
  distribution.
  
\item \textbf{Entropy-Threshold Policy (ET)}: we compute the entropy of all
  policies at state $s_t$ and find the sub-policy $k^*$ with minimum entropy:
  \begin{eqnarray}
    \mathcal{H}_{k} &=& - \ln\frac{1}{|A|} \sum_{a} \pi_k(a | s_t) \ln \pi_k(a | s_t) \\
    k^{*} &=& \underset{k=1,\ldots,K}{\mathrm{argmin}} \mathcal{H}_k \label{eq:min-entropy},
  \end{eqnarray}
  Where $|A|$ is the cardinality of the state space. 
  Then, if $\mathcal{H}_{k^*}$ is less then $\mathcal{H}_{0}$ plus threshold
  $\epsilon$, we perform the action following the sub-policy, otherwise we
  perform the action following the main policy:
  \begin{equation}
    \pi_{f}(a | s_t) = 
    \begin{cases}
      \pi_{k^*}(a | s_t) & \text{if} \quad \mathcal{H}_{k^*} < \mathcal{H}_{0} + \epsilon \\
      \pi_0(a | s_t) & \text{otherwise}
    \end{cases}
  \end{equation}
  
\item \textbf{Entropy-Weighted Mixture Policy (EW)}: the resulting policy is a weighted average of the main policy and the minimum-entropy sub-policy identified using equation~\ref{eq:min-entropy}:
  \begin{equation}
    \pi_{f}(a|s_t) =  \mathcal{H}_{k^*} \times \pi_0(a|s_t) + (1 - \mathcal{H}_{k^*}) \times \pi_{k^*}(a|s_t).
  \end{equation}
  This is a combination of MP and ET.
\end{itemize}

\section{Experiments}


We performed experiments on two different environments 
to compare the
policy fusion methods. In the first experiment, we combine two independent
policies trained with hard-coded reward functions in order to understand the
performance of each fusion method. For this, we used the MiniWorld
\cite{miniworld} environment, a minimalist 3D interior environment simulator for
reinforcement learning and robotics research. In the second set of experiments,
we used the DeepCrawl environment \cite{sesto19} to evaluate how policy fusion
methods can be used in combination with Inverse Reinforcement Learning (IRL).

\subsection{Results on MiniWorld}

For our first experiment we use the {\tt PickUpObjs} variant of MiniWorld. In
this environment, there is a single large room in which the agent must collect
objects of two types: red boxes and green balls. A maximum of 5 objects are
spawned in random positions. The observation space of this environment is a
single RGB image of size $(80, 60, 3)$.

We train two policies with different, hard-coded reward functions.
Our main policy $\pi_0$ is trained to collect all the red boxes,
with reward:
\begin{equation}
    R_0 = +1 \; \; \text{for collecting a red box}.
\end{equation}
A single sub-policy $\pi_1$ is then learned with the aim of collecting all the 
green balls:
\begin{equation}
    R_1 = +1 \; \; \text{for collecting a green ball}.
\end{equation}

\begin{figure}
    \centering
    \includegraphics[width=\plotwidth\textwidth]{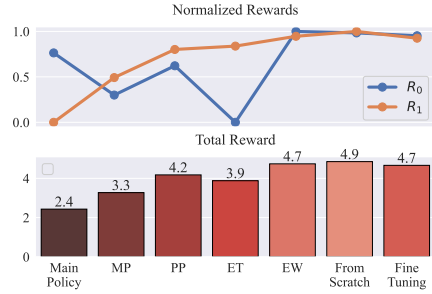} 
    \caption{ Results of the different fusion methods in the MiniWorld
      environment. The plot shows the normalized rewards $R_0$ and $R_1$ using
      the methods described in section \ref{sec:methods}. The plot below shows
      the total reward $R_0 + R_1$ achieved by the respective combined policies.
      Numbers are averages over 1,000 episodes.}
	\label{fig:results_miniworld}
\end{figure}

\begin{figure*}
    \begin{center}
    \scalebox{1.0}{
    \begin{tabular}{ccc}
        \includegraphics[width=\plotwidth\textwidth]{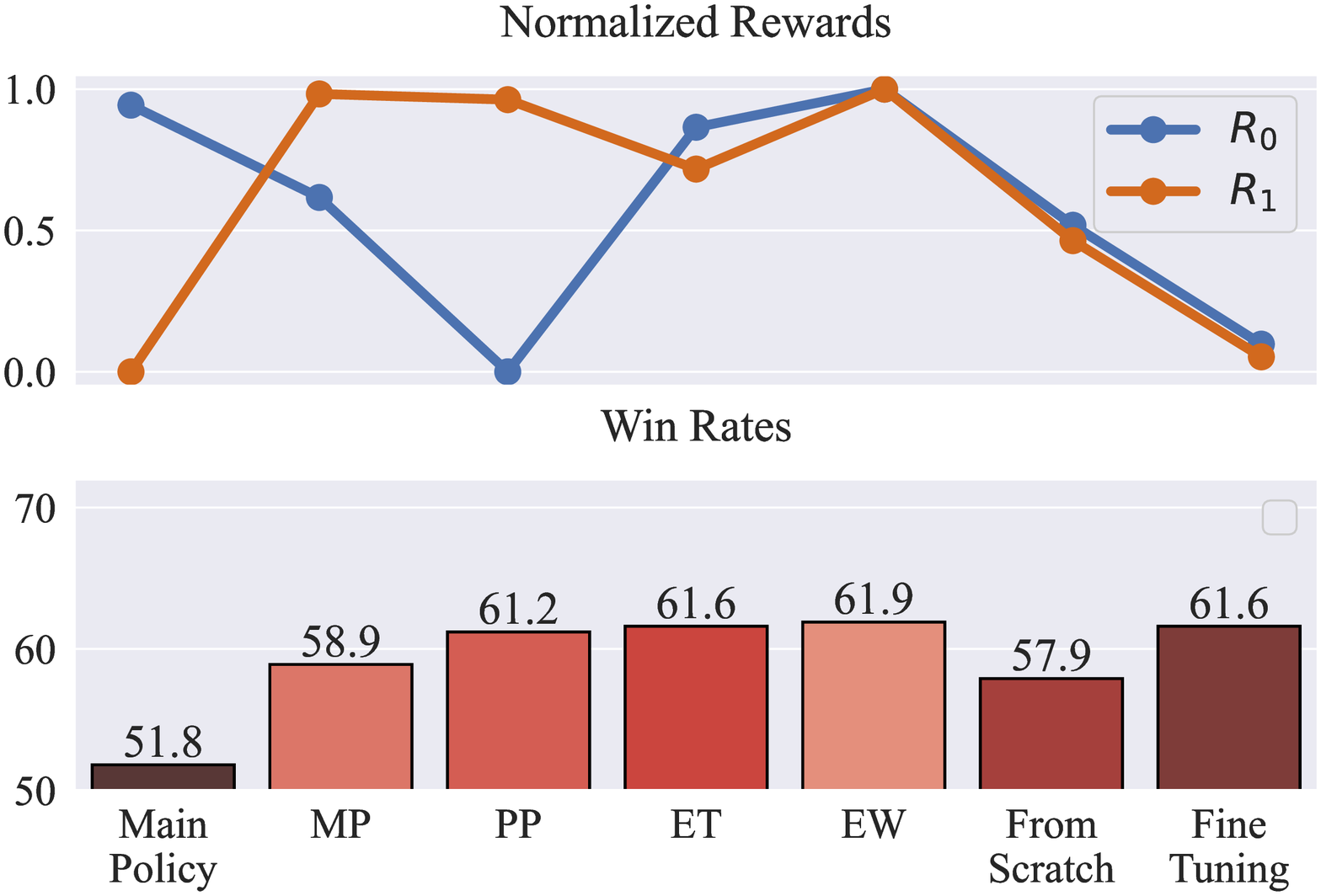}  &
        \includegraphics[width=\plotwidth\textwidth]{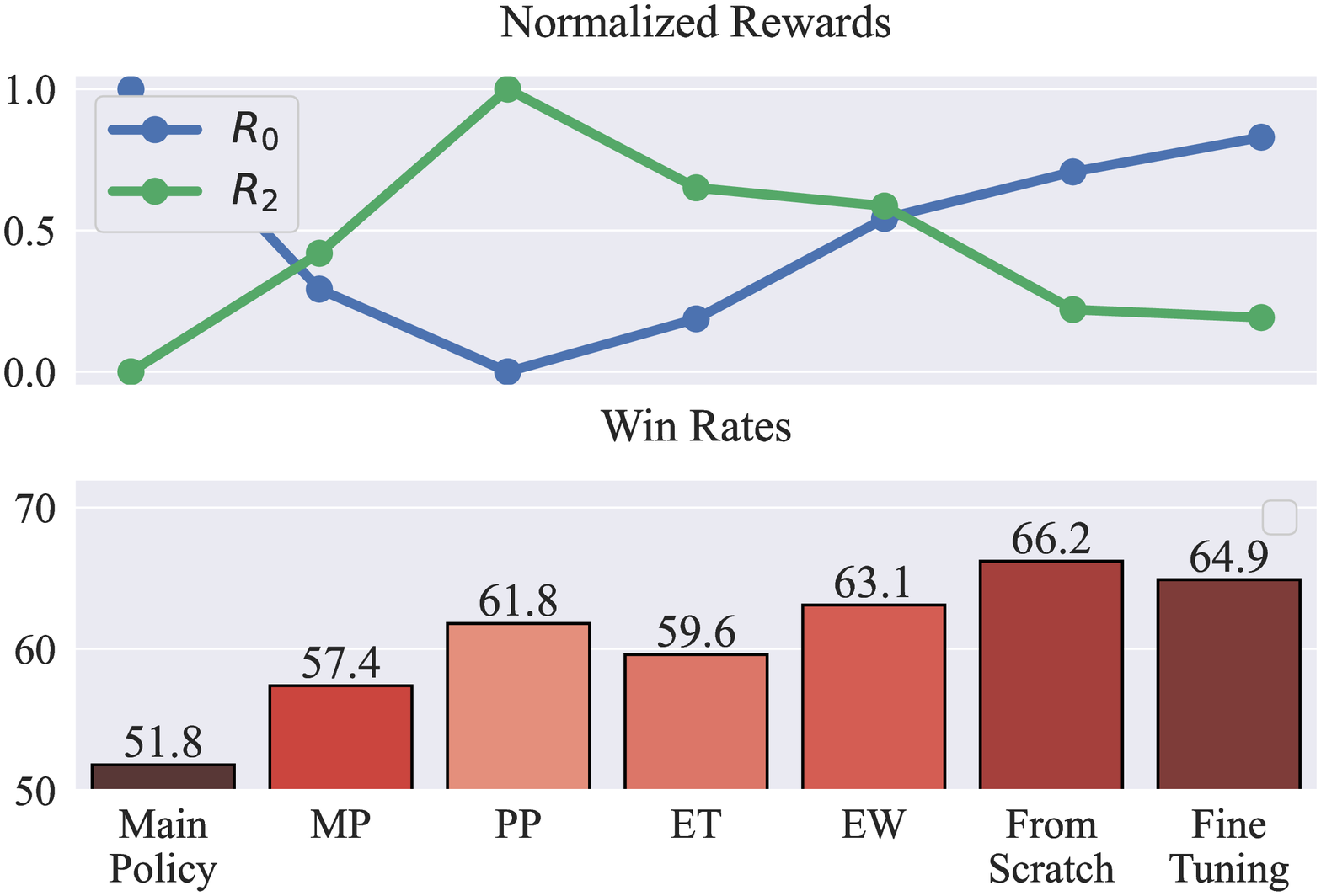} &
        \includegraphics[width=\plotwidth\textwidth]{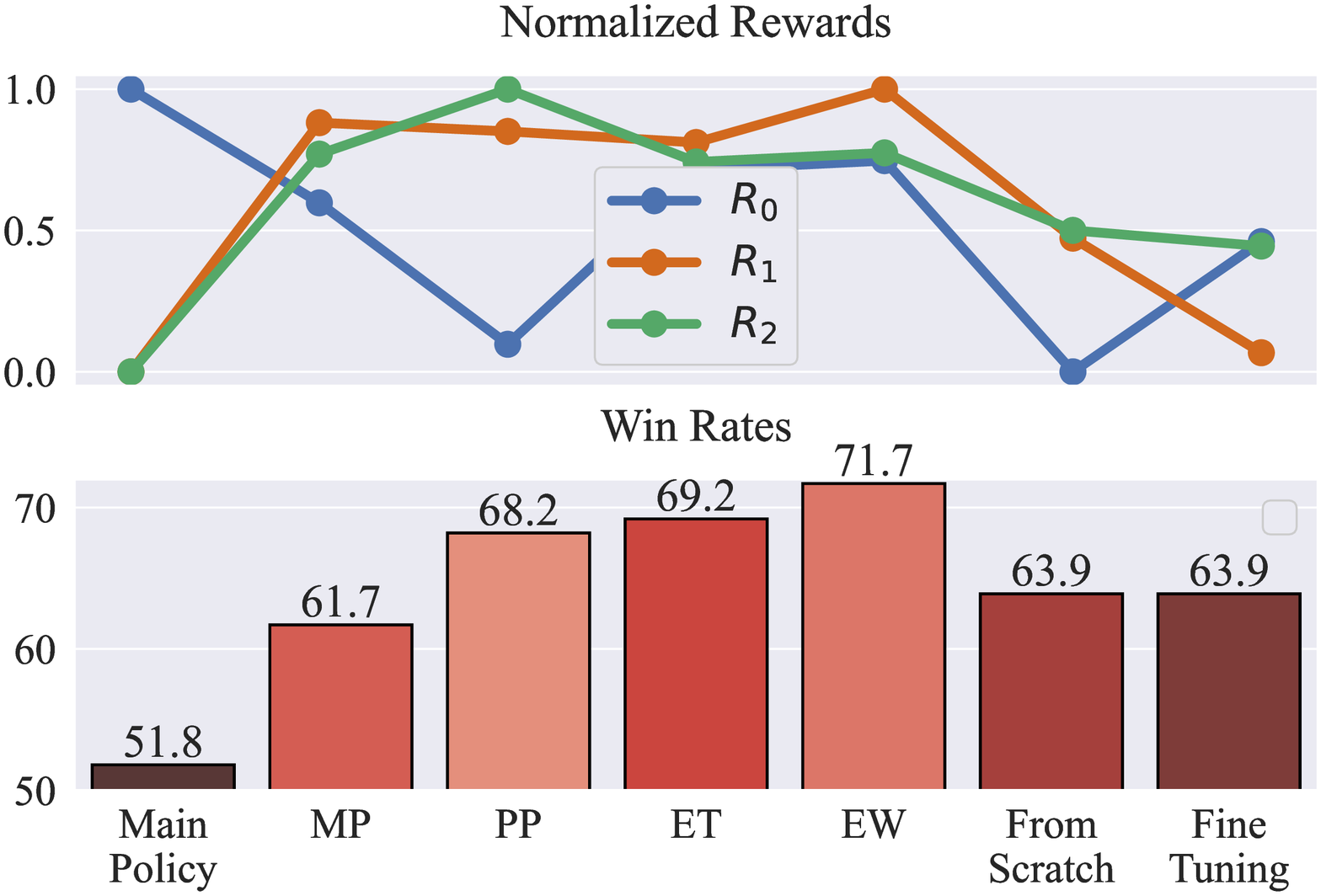} \\
        (a) & (b) & (c)\\
        
    \end{tabular}
    }
    \end{center}
    \caption{Results for the \textit{enhancing performance} use-case. In all
      figures, the top plot shows the normalized rewards for each fusion method:
      $R_0$ refers to the original environment reward, $R_1$ refers to the
      reward used for training sub-policy $\pi_1$, and $R_2$ to the reward used
      for training sub-policy $\pi_2$. The bottom plot shows the win rate of the
      combined agents versus the base agent $\pi_0$. (a) Results for the
      combination $\pi_0 + \pi_1$. (b) Results for $\pi_0 + \pi_2$. (c) Results
      for $\pi_0 + \pi_1 + \pi_2$. Numbers are averages
      over 1,000 episodes.}
    \label{fig:results_competitive}
\end{figure*}

The two policies use the same network architecture, consisting of three, stride
2 convolutional layers of size $5$, $3$ and $3$, respectively, and with 32, 32,
and 64 channels, respectively. These are followed by two fully connected layers
of size $256$ and a final fully connected layer followed by a softmax to
represent the action distribution. The action space consists of $5$ discrete
actions: turn left, turn right, move forward, move backward and pick up.

After training the two policies with Proximal Policy Optimization (PPO)
\cite{ppo}, we combine them using the methods proposed in section
\ref{sec:methods}. As baselines, we also train a policy from scratch using the
combination of $R_0 + R_1$, as well as a policy fine-tuned for the combined
reward $R_0 + R_1$ after being trained first with only $R_0$. Whereas the fusion
methods do not involve further learning, the two baseline policies represent
``upper bounds'' which are trained to optimize the combined reward.

Figure \ref{fig:results_miniworld} shows the results of these experiments. As
the plots illustrate, each policy fusion method improves upon the overall
performance of the agent with respect to the combined reward. However, our
proposed fusion approach EW outperforms all other methods and achieves the same
performance level of the policies trained from scratch or fine-tuned.

This first experiment shows that our proposed policy fusion
methods can indeed combine different policies to achieve more complex behavior.
The results indicate that, while all yield some improvement, the EW method
is by far the best overall, followed by PP. This is a trend we observed in
all subsequent experiments.\footnote{Code to replicate these experiments is
at \url{https://tinyurl.com/fusion-rl}}


\subsection{Results on DeepCrawl}
DeepCrawl is a Roguelike game built for studying the applicability of DRL
techniques in video game development~\cite{sesto19}. The visible environment at
any time is a grid of $10 \times 10$ tiles. Each tile can contain the agent, an
opponent, an impassable object, or collectible loot. The structure of the map
and object locations are procedurally generated at the beginning of each
episode. Collectible loot and actors have attributes whose values are randomly
chosen as well. The action space consists of 19 different discrete actions: $8$
movement actions, $8$ ranged attack actions, $2$ magic actions and $1$ loot
action. The environment reward function is defined as:
\begin{equation}
    R_0= -0.01 + 
    \begin{cases}
        -0.1 & \text{for an impossible move} \\
        +10.0*\mathrm{HP} & \text{for the win}
    \end{cases},
    \label{eq:reward}
\end{equation}
where HP refers to the normalized agent HP remaining at the moment of defeating
an opponent. Agents are trained against an enemy that always makes random moves.
For a complete description of the environment and the training set-up, refer to
the original paper \cite{sesto19}.

As before, we used PPO to train agents in this environment, and they all use the
architecture proposed in \cite{transformers}. We train an agent with the
hard-coded reward function in equation \ref{eq:reward} as the main policy
$\pi_0$. The training of this agent reflects the ``Ranger'' training described
in \cite{transformers}. Subsequently, we train one or more sub-policies and
merge those with the main policy. Each of the following sub-policies are trained
with DE-AIRL, using the reward approximators of \cite{deairl}. The majority of
demonstrations for DE-AIRL come from a human expert. In the next section we
describe the experiments we conducted, each of which are instances of a use-case
outlined in section \ref{sec:intro}.

\begin{figure*}
    \begin{center}
    \scalebox{1.0}{
    \begin{tabular}{ccc}
        \includegraphics[width=\plotwidth\textwidth]{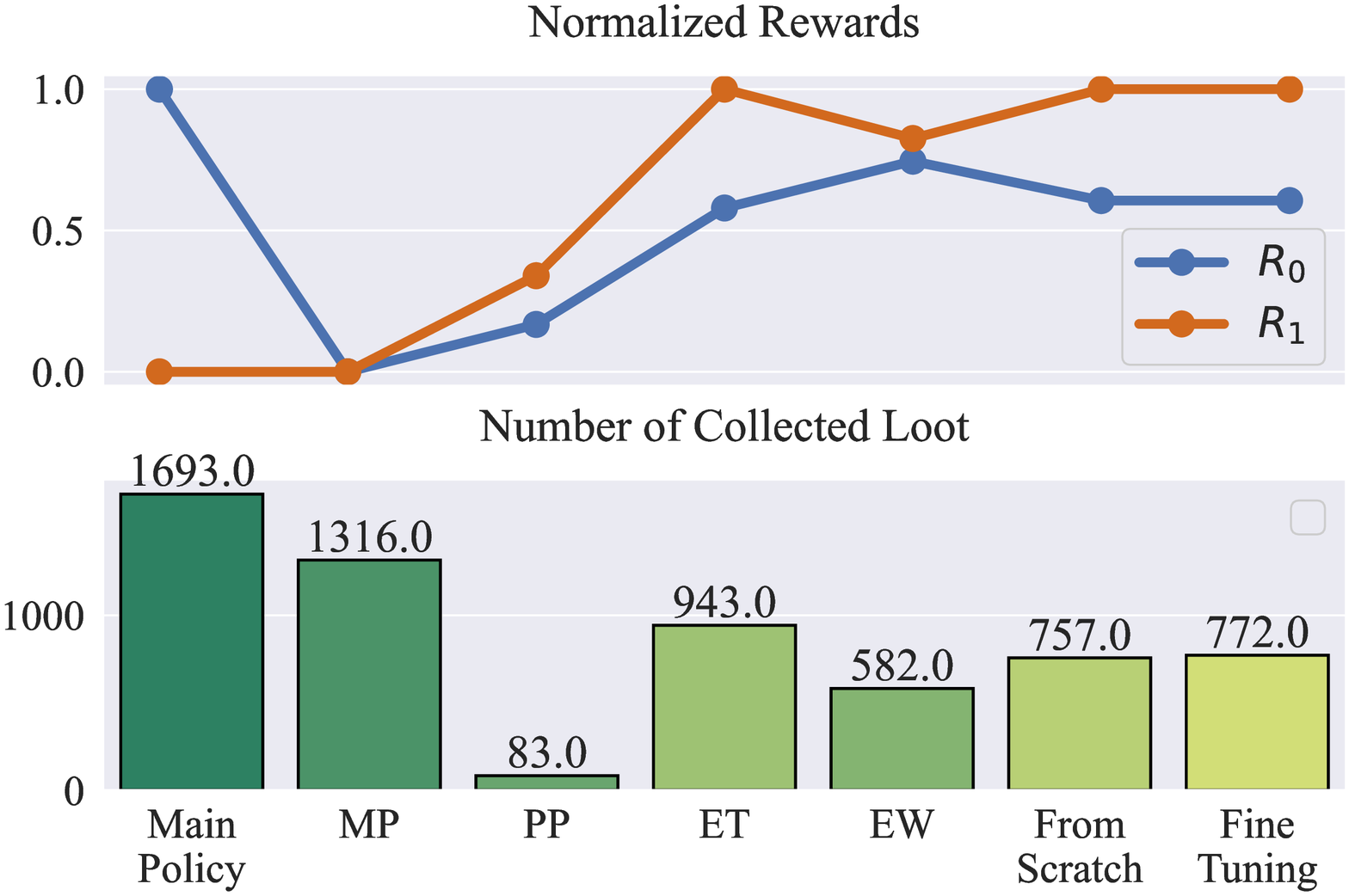} &
        \includegraphics[width=\plotwidth\textwidth]{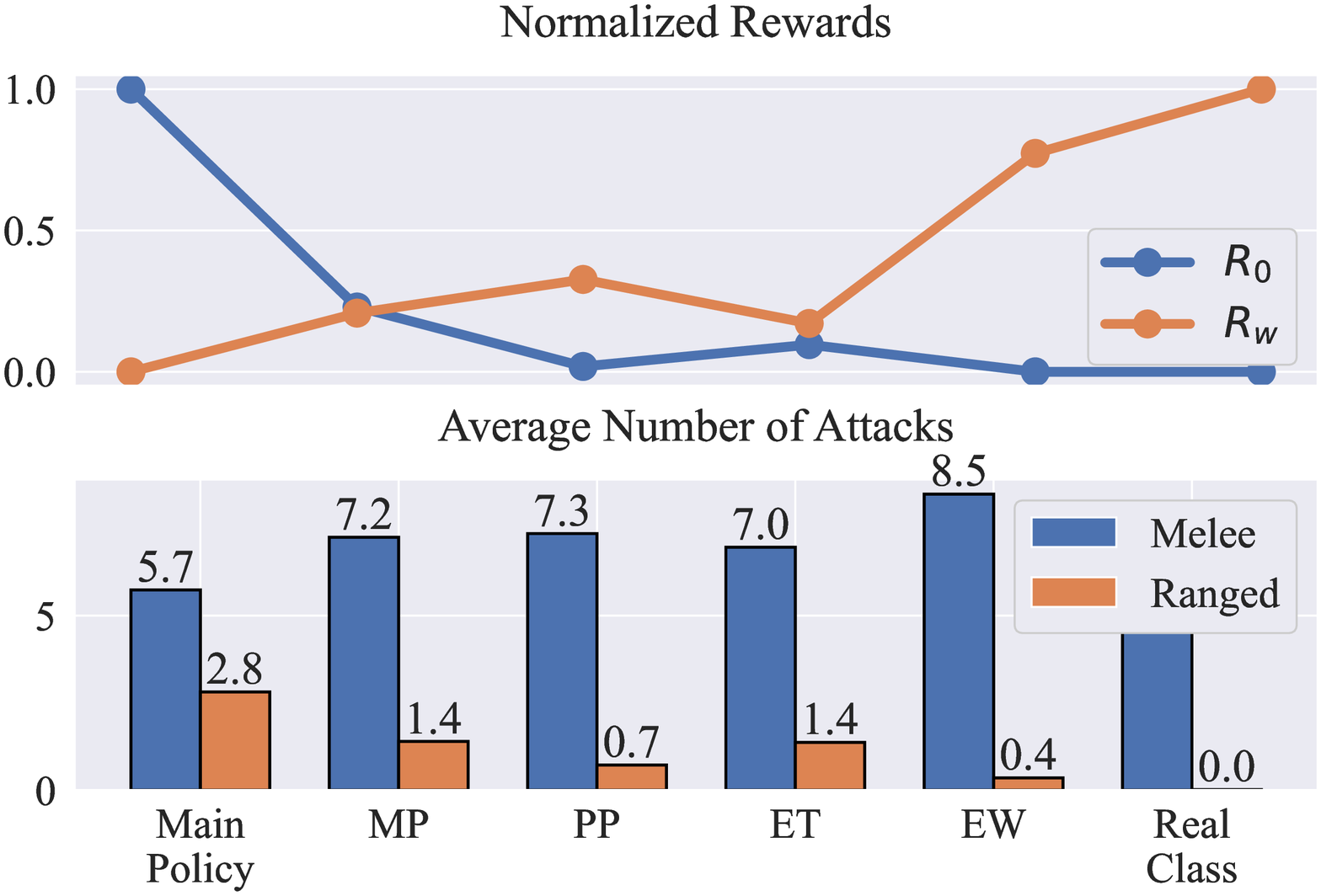} &
        \includegraphics[width=\plotwidth\textwidth]{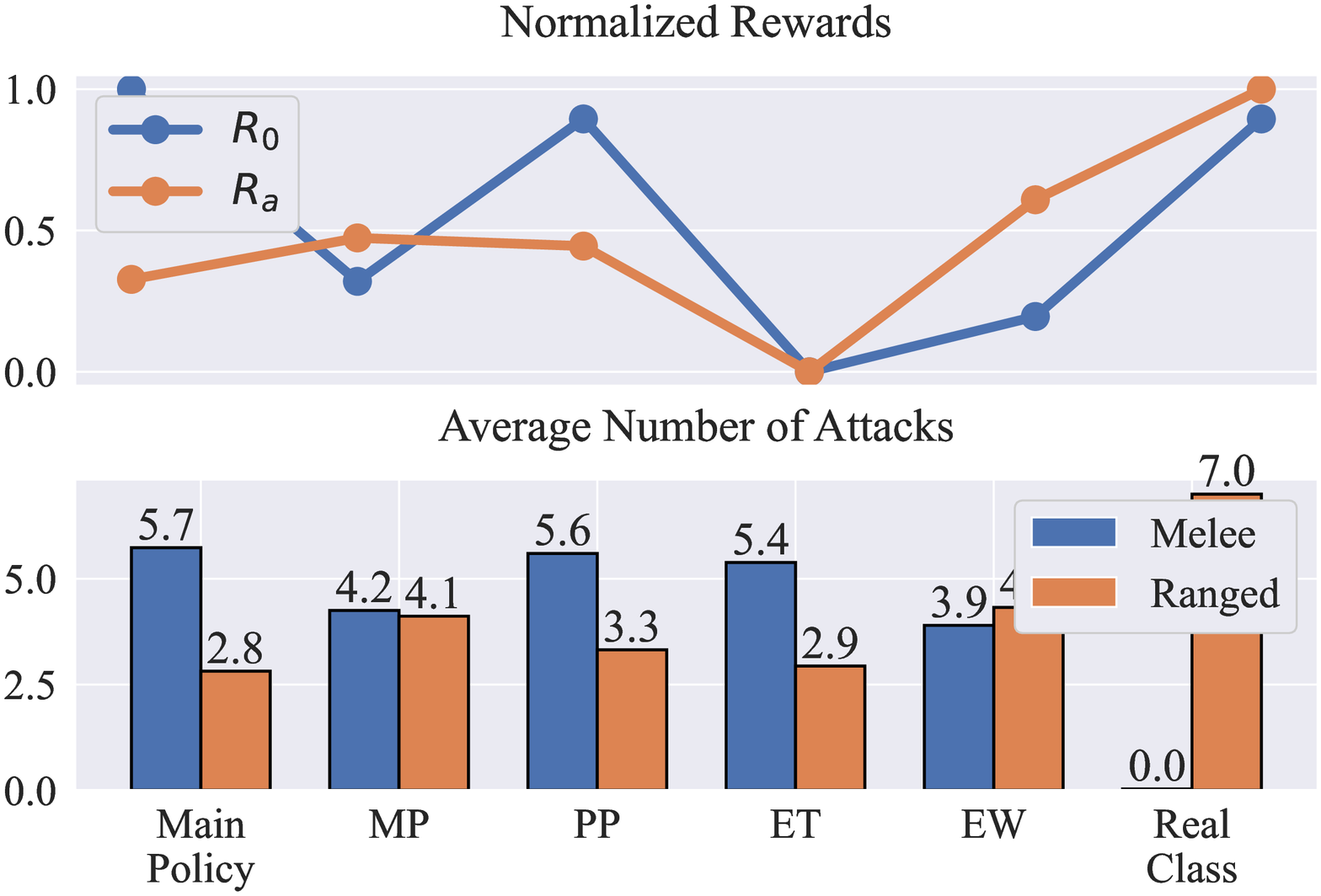}  \\
        (a) & (b) & (c) 
    \end{tabular}
    }
    \end{center}
    \caption{Results for the \textit{adding style} use-case. In all figures, the
      top plot shows the individual normalized rewards for the different fusion
      methods: $R_0$ refers to the original environment reward, and $R_*$ refers
      to the reward used to train sub-policy $\pi_*$. The bottom plot
      shows some qualitative statistics which define the ``style'' of the agent.
      (a) Combination $\pi_0 + \pi_1$, where $\pi_1$ is a
      sub-policy trained to avoid loot in the map. (b) Combination
      $\pi_0 + \pi_w$, where $\pi_w$ is a sub-policy trained to act like a
      ``warrior''. (c) Combination $\pi_0 + \pi_a$, where $\pi_a$ is a
      sub-policy trained to act like an ``archer''. Numbers are 
      averages over 1,000 episodes.}
    \label{fig:results_classes}
\end{figure*}

\minisection{Enhancing performance.}
\label{sec:increase}
Suppose that training with the hard-coded reward function does not result in
competitive agents. For example, they do not use certain objects in the map
which designers believe would improve their win rate. This can be caused by a
badly-designed reward function or a sub-optimal training setup.

To explore this use case, we created two different sub-policies $\pi_1$ and
$\pi_2$ for our first experiment. The first is trained with DE-AIRL to collect
and use a specitic object in the map, while the second is trained from
demonstrations to get all loot which increases agent statistics. We then combine
$\pi_0 + \pi_1$ and $\pi_0 + \pi_2$, and let the combined agents fight against
the base agent $\pi_0$. Pitting agents against the base agent allows us to have
some quantitative measure about the different policy fusion methods, as our aim
here is to create agents that increase their win rate.

As figure \ref{fig:results_competitive}a and \ref{fig:results_competitive}b
show, combining the main policy with $\pi_1$ and $\pi_2$ increases the win rate of
the main agent for all fusion methods, with EW being the best, followed by
PP.
Moreover, all methods decrease the original reward function of the environment.
This indicates that the reward is probably not optimally designed in order to
yield the most competitive agents (in terms of win rate).

The experiment shows that combining policies can increase the win rate
of an agent by more than $10\%$. Among the methods, EW is the one that decreases
the original reward the least while at the same time improving the sub-policy
reward. In the $\pi_2$ experiment, the EW method even outperforms training from
scratch and fine-tuning. This is probably due to the difficulty of combining a
hard-coded reward function with a learned one, since these rewards
have different scales.

Since both sub-policies individually already increase the competitiveness of the
main agent, we also tried to combine them all: $\pi_0 + \pi_1 + \pi_2$. As we
expected, figure \ref{fig:results_competitive}c shows that the combination of
all three policies using EW results in an even more competitive agent which wins
more than $70\%$ of the games against the base policy $\pi_0$. Furthermore, EW
clearly outperforms the other methods, including training from scratch and
fine-tuning.

\minisection{Adding style.} 
\label{sec:style}
This use case is about training NPCs that exhibit certain behavioral styles
specified and controlled by developers. For example, designers may want an agent
to act more ``sneakily'' or more ``aggressively''. It is very difficult to
computationally specify subjective styles, and expect that the main agent
$\pi_0$ may likely not reflect the qualitative behavior a designer wants. Since
we are not interested in the \emph{competitiveness} of the agent here, but
rather being able to control the behavioral aesthetics of NPCs, we conduct a
qualitative analysis of agent behavior.

We first train a sub-policy $\pi_1$ with DE-AIRL which \emph{avoids} loot in the
map while fighting against the opponent. This is an interesting experiment for
two reason: on the one hand, we add a new style to the main behavior, while, on
the other hand, we are trying to limit the agent avoid a behavior it has already
learned during training of $\pi_0$. Figure \ref{fig:results_classes}a shows that
EW method continues to outperform the others, as it decreases the main reward
the least while simultaneously augmenting it with the style demonstrated by the
designers. Other interesting observations here are that MP does not work at all,
while the PP method does add the intended style to the behavior but deviates a
lot from the main policy.


For the second experiment in this use-case changed the combat style of $\pi_0$
to emulate the style of the other two agent classes in \cite{sesto19} -- Warrior
and Archer. These two classes are distinguished from each other by the type of
attacks they perform in combat: the Warrior relies on melee attacks, while the
Archer only on ranged attacks. In contrast, the Ranger, our main policy,
performs both melee and ranged attacks. We train Warier and Archer sub-policies
$\pi_w$ and $\pi_a$ from demonstrations provided by the pre-trained agents and
then combine them with the main policy: $\pi_0 + \pi_w$ and $\pi_0 + \pi_a$. The
results are shown in figures \ref{fig:results_classes}b and
\ref{fig:results_classes}c. Indeed, with $\pi_0 + \pi_w$ we are able to modify
the Ranger behavior act as a Warrior. With $\pi_0 + \pi_a$ we can change the
behavioral style of the Ranger \emph{towards} the Archer, i.e., to perform more
ranged than melee attacks, but the combination does not perfectly emulate the
Archer behavior. Again, in both cases the EW method outperforms all other fusion
methods.

\minisection{Adapting to new features.} 
\label{sec:adapt}
suppose an agent $\pi_0$ was trained on a certain version of the environment,
$E_0$. Later, after design decisions, some aspect of the game has changed. For
example, maybe a new usable item in the map was added, to arrive at a new
version of the environment $E_1$. At this point, designers require an agent
$\pi_1$ which is aware of and able to properly exploit this new feature. They
have two choices: re-train from scratch all previously-trained agents, or train
a sub-policy in order to teach it how to use only the new object, and then
augment $\pi_0$ with this sub-policy.

\begin{figure*}
\begin{center}
\scalebox{1.0}{
    \begin{tabular}{cc}
        \includegraphics[width=\plotwidth\textwidth]{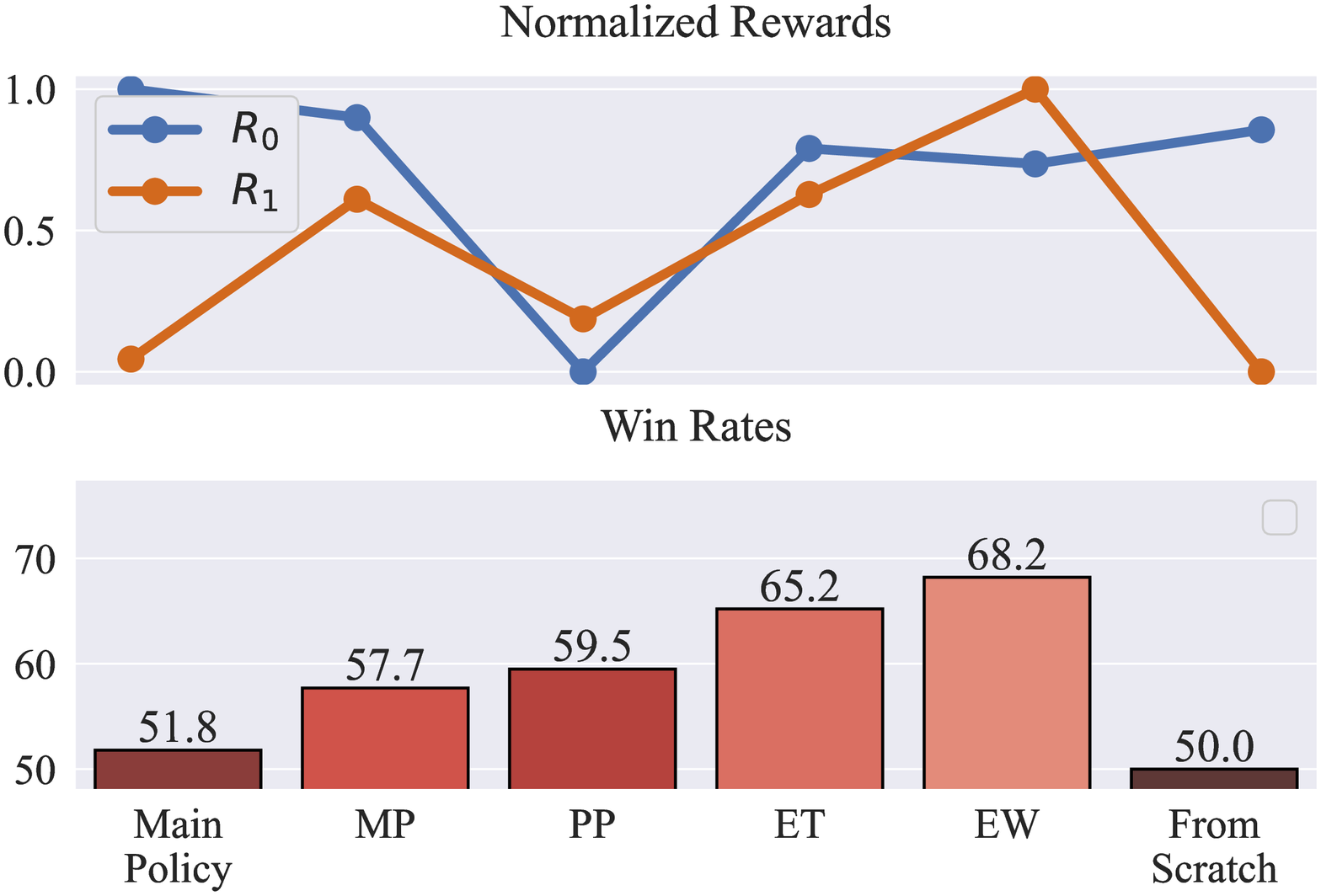} &
        \includegraphics[width=\plotwidth\textwidth]{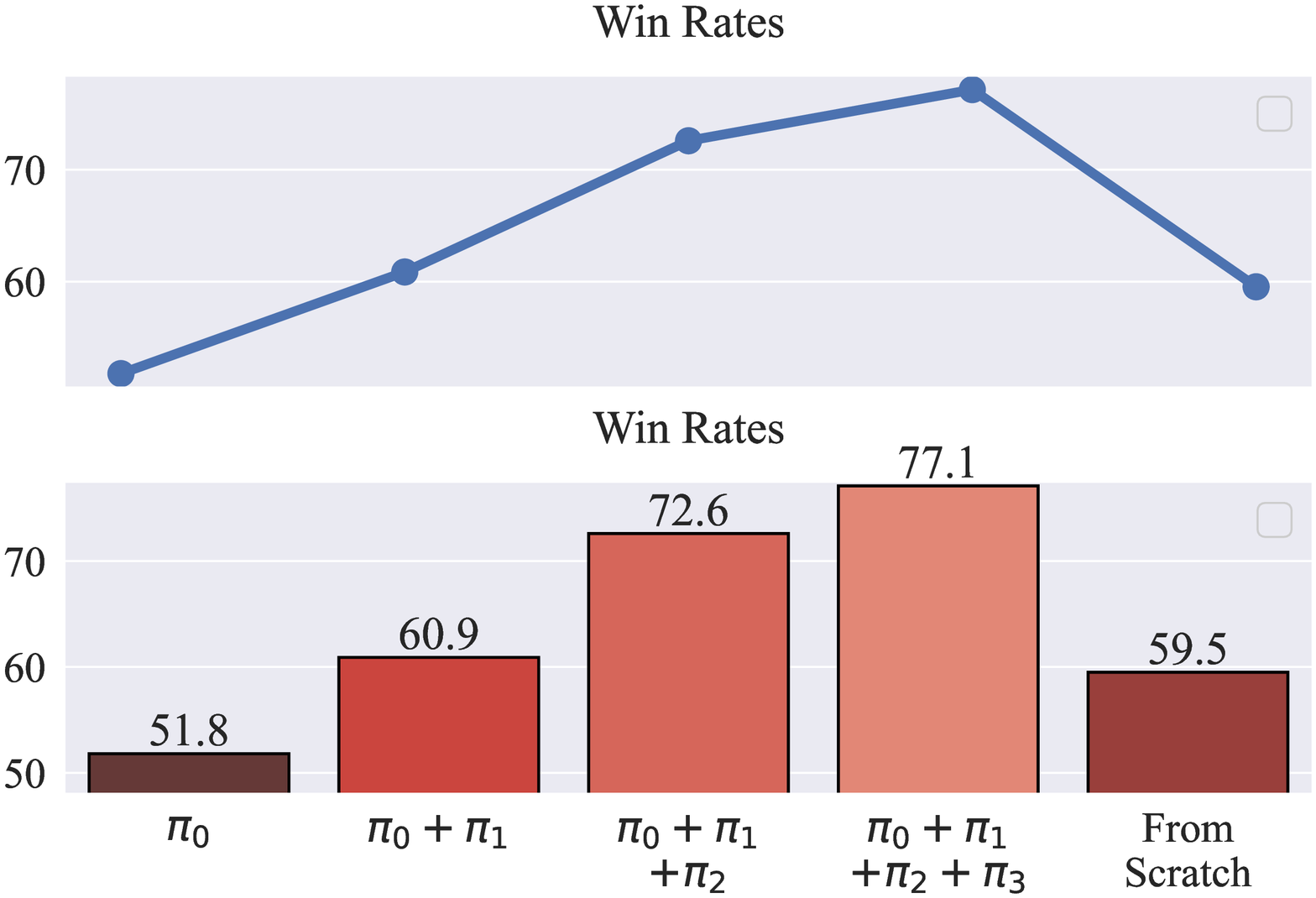}  \\
        (a) & (b)
    \end{tabular}
    }
    \end{center}
    \caption{Results for the \textit{adapting to new features} use-case. (a) The
      top plot shows normalized rewards for the different policy fusion methods:
      $R_0$ refers to the original environment reward and $R_1$ refers to the
      reward used for training sub-policy $\pi_1$. The bottom plot shows the win
      rate of the combined agent versus the base agent. All experiments use the
      $E_1$ version of the game. (b) Both plots show the win rate of
      the combined agent versus the plain agent with different combinations of
      sub-policies with $\pi_0$. Numbers averages over 1,000 episodes.}
    \label{fig:results_adapt}
\end{figure*}

For our first experiment in this use-case, we added a new usable object to the
game. We designed this object so that an NPC using it will have an advantage
agents that do not. This way, use of the new object will be reflected in the win
rate of $\pi_1$ over $\pi_0$. We train a sub-policy to exploit this feature and
combine it with $\pi_0$ using policy fusion. This requires an ``adapter'' which
hides the new object from $\pi_0$, so to avoid confusion of encountering a new
object never seen during training. As baseline, we train an agent from scratch
in $E_1$ but with the original DeepCrawl reward defined in equation
\ref{eq:reward}. Figure \ref{fig:results_adapt}a shows the results: the combined
agent is perfectly capable of using the new object, with EW outperforming all
other fusion methods. An interesting observation here is that training from
scratch does not learn to make use of the new feature: it just ignores the
object and reaches the same level of performance as $\pi_0$. This demonstrates
that training from scratch in the face of design changes is not a trivial task:
we may need to tune hyperparameters or even adapt the reward function to train
an agent able to incorporate the new feature into its behavior. But, as we
mentioned before, engineering a well-designed reward function for very complex
behavior is not easy, and training from scratch takes more time than training
only a sub-policy. Our approach enables quick and efficient adaptation to new
features in a game environments, which facilitates testing without having to
wait for agents to re-train.

Our final experiment simulates multiple, iterative design changes: since we know
that fusion methods can adapt to a single change in the environment dynamics,
what if we add two more features? For this, we first added a new usable object
which increases the statistics of agents that collect it, and then we added an
instant death tile. Clearly, an agent that adapts to these changes will have an
advantage. For evaluation we compare two agents in an environment with both new
features present. The first agent $\pi_0$ was trained before these features were
added. The second agent is $\pi_0$ plus all the sub-policies trained how to use
the new features. The baseline is an agent re-trained from scratch on the
environment with both changes. Since we have established that EW is the best of
the considered fusion methods, we only used this technique here. Figure
\ref{fig:results_adapt}b shows the results for different combinations of
sub-policies and $\pi_0$. Indeed, we can even combine even all $4$ policies and
achieve better results than the re-training baseline. We believe that after some
hyperparameter tuning, the baseline will likely be able to perform better than
our method. However, this would come at a very high cost in term of human effort
and time.

\subsection{Training Times}
\label{sec:training_times}
Taking the last experiment of \ref{sec:adapt} as illustrative example, training
the main policy takes about $66$ hours, while training a single sub-policy
requires about $6$ hours. Using policy fusion thus requires only an additional 6
hours for each adaptation, whereas it takes another $66$ hours per adaptation to
re-train main policy from scratch. Our method is about $10$ times faster than
the standard approach. All training was performed on a NVIDIA RTX 2080 SUPER GPU
with 8GB RAM.



\section{Conclusions}
Training intelligent agents in complex environments using Deep Reinforcement
Learning is difficult and time-consuming, and moreover requires specialized
knowledge of both the domain and state-of-the-art deep learning techniques. In
this paper we presented several policy fusion methods that can \emph{combine}
policies with the aim of adapting or modifying behavior in the face of game
design changes -- all without requiring retraining of agents to cope with these
changes. Our experiments clearly show that the Entropy-Weighted Mixture (EW)
fusion technique significantly outperforms the others methods, and in some cases
even surpasses the agent re-trained from scratch or fine-tuned using the
combined reward function. Of all methods, EW strikes the best balance of
maintaining the original policy behavior while simultaneously augmenting it with
the sub-policy's new ``style''.

Our experiments also showed how these fusion methods can be used in combination
with Inverse Reinforcement Learning to create varied and complex behavior
without defining new reward functions, which contrasts to the prevailing
perception that IRL is difficult to exploit in high-dimensional state
spaces~\cite{elettronicarts}.

Creating credible and meaningful NPC behavior is a difficult task. We think that
our work is a first step towards more controlled behavior design by blending
interpretable fusion and DRL training. In future work, we are planning to
investigate more sophisticated policy fusion methods which give more control
over agent behavior. For instance, we are currently studying ways to learn which
policy is the best to use in a particular state. Finally, we are interested in
comparing multi-objective techniques for specifying attitudes (e.g.
\cite{microsoftblog}) to our fusion methods.




\bibliographystyle{IEEEtran}
\bibliography{main}

\end{document}